# Intelligent Image Sensing for Crime Analysis: A ML Approach towards Enhanced Violence Detection and Investigation

Aritra Dutta, Pushpita Boral, Dr. G Suseela

Department of Networking and Communications,

School of Computing,

Faculty of Engineering and Technology,

SRM Institute of Science and Technology,

Kattankulathur, Tamil Nadu, 603203, India

ad9382@srmist.edu.in

*Abstract*- **The increasing global crime rate, coupled with substantial human and property losses, highlights the limitations of traditional surveillance methods in promptly detecting diverse and unexpected acts of violence. Addressing this pressing need for automatic violence detection, we leverage Machine Learning to detect and categorize violent events in video streams. This paper introduces a comprehensive framework for violence detection and classification, employing Supervised Learning for both binary and multi-class violence classification. The detection model relies on 3D Convolutional Neural Networks, while the classification model utilizes the separable convolutional 3D model for feature extraction and bidirectional LSTM for temporal processing. Training is conducted on a diverse customized datasets with frame-level annotations, incorporating videos from surveillance cameras, human recordings, hockey fight, sohas and wvd dataset across various platforms. Additionally, a camera module integrated with raspberry pi is used to capture live video feed, which is sent to the ML model for processing. Thus, demonstrating improved performance in terms of computational resource efficiency and accuracy.**

*Index Terms*- Panoramic Imaging, Violence Detection, Crime Analysis, Image sensing, Raspberry Pi, Low-cost Computing, Surveillance, Automated Surveillance systems, Computer Vision

## I. INTRODUCTION

Installation of CCTV cameras has become a necessity nowadays almost everywhere including shopping malls, theatres, houses and restaurants. This is due to the increase in violent activities all over the world and the importance of monitoring it to reduce human and property loss. But simply keeping an eye is not enough, rather it is very crucial to recognize and classify it timely. For this, [3] have developed a system based on transfer learning to finely classify violent behavior using pre trained Movie and hockey datasets. Further, The paper utilizes GoogleNet, a 22-layer deep neural network, as a pre-trained source for transfer learning experiments. GoogleNet is chosen for its efficiency, having 12 times fewer parameters than AlexNet, and is trained on the ImageNet dataset with 15 million annotated images across 1000 categories. In the experiments, the last classification layer is modified to distinguish between violent/fight actions and non-fight actions using 2 classes from the Hockey and Movies dataset. Hockey dataset primarily consists of 1000 recorded hockey game videos while Movie dataset contains 200 dynamic video clippings which include scenes from different scenarios including both violent and non-violent behavior.

Human activity recognition has been studied very deeply in recent years and is of interest to the thriving research community. It is highly indulging due to the various diverse aspects it brings to the table, but also has a lot of real-time challenges to be addressed. This task can be classified into two broad domains namely human posture detection and human action recognition. A unique multi-task deep learning method for action recognition and estimation of human morphology is presented in the paper in [1]. The proposed 3D pose method achieves high precision with low-resolution feature maps, avoiding the need for costly volumetric heat maps by predicting specialized depth maps for body joints. The CNN architecture, combined with previous posture detection enhances efficient multi-scale individual objects and action management. The model is trainable with a mix of 2D and 3D data, showcasing significant improvements in 3D pose estimation. Simultaneous training with single frames and video clips is seamless. The carefully designed architecture effectively addresses the challenge of multitasking human poses and action recognition, outperforming separate task learning. Joint learning of human poses consistently enhances action recognition, and the model is highly scalable, offering flexibility in cutting at mutiple levels for predicting actions.

Based on such findings and observations, researchers went on to explore human pose study to a greater extent. This paper [2] introduces a machine for 3D human posture that effectively integrates long-range spatio-temporal dependencies and three dimensional configuration in a comprehensive and suggested way. The model is improved by introducing a unique correction mechanism which is self-supervised, involving two different learning tasks: 2D-to-3D conversion of poses and 3D-to-2D pose estimation. This correction mechanism maintains geometric regularity among 2D manipulation of 3D objects and calculated 2D poses, allowing the model to further procress intermediate pose estimation bidirectionally using the predicted 2D human pose. Importantly, the given correction mechanism bridges the gap between three dimensional and two dimensional human poses, enabling the implementation of external 2D human pose data without the need for extra three dimensional annotations. Further research will focus on extending the correction mechanism which trains itself for time referenced relationship modeling in sequential human based activities, such as different physical actions and recognizing them. Additionally, newly designed supervision targets will be designed to incorporate various three-dimensional geometric attributes for cost-effective model building.

To integrate such solutions with real-time cases, it was important that devices should be movable and can capture features at any point in the space. The paper [6] introduces Wi-Mose, a novel device to determine movable 3D objects utilizing standard WiFi gadgets. The system is divided into three main components: dataset acquisition, processing of data, and estimation of poses. The data collection process involves the usage of dual receivers, a transmitter, and a compact refracting camera to gather video frames and Channel State Information (CSI). In the initial stage of processing data, unrefined CSI items are converted into CSI images, and consecutive video frames are transformed into human key-point coordinates for supervised learning. The position estimation element simplifies the process of reconstructing three dimensional human posture skeletons by extracting characteristics from CSI pictures and converting them into key-point values. By constructing CSI images, the system allows a neural network for deriving pose based specifications that are independent of location. The designed neural network converts these details into salient values. The results of the experiment demonstrate that Wi-Mose achieves improved accuracy, with 29.7mm and 37.8mm P-MPJPE in Line-of-Sight (LoS) and Non-Line-of-Sight (NLoS) scenarios, representing a 21% and 10% enhancement compared to the baseline. Future work aims to extend Wi-Mose's application to various environments for increased accuracy of 3D movable human pose calculation.

## II. RELATED WORK

It is important to analyze the various techniques used so far for crime scene analysis through different parameters and conditions leading to significant advancement in this field. This task can be broken down into two broad spectrums, namely action recognition and classification.

**2.1.** *3D Pose Estimation*

Diogo C. Luzivon [1] aims to develop a system which detects poses and recognizes action of humans in 2D and 3D architecture simultaneously. The model is trained using both
*2D* images and video clips to attain improved 3D pose estimation. It utilizes MPII Human Pose Dataset, which is a two dimensional dataset with 25,000 images from YouTube, annotated for sixteen joints in human body. The Human3.6M is a three dimensional dataset capturing eleven entities performing seventeen different activities at the same time with four cameras, annotated with elevated three dimensional poses for 17 joints. Penn Action is a 2D dataset for action recognition comprising of 2,326 different videos of sports-related activities, which are marked manually for thirteen human body joints. The NTU RGB+D consists of 56,000 Full HD videos, featuring around sixty actions conducted by forty actors, and registered by three cameras, including color videos, depth maps, and 3D Kinect poses. The model built using these datasets is much more accurate and scalable as compared to previous approaches of pose detection and action recognition as separate tasks.

**2.2.** *Transfer Learning*

Violence detection is of utmost importance to develop video surveillance cameras. To do so, researchers [3] have trained a deep learning CNN model utilizing the Hockey and movie dataset for fight action recognition using 10-fold cross validation technique. This system achieved around 99% accuracy for both the hockey and movie datasets to detect crime scenes.

The system in [5] utilizes a Convolutional Neural Network (CNN) based on the VGG-16 architecture as a system for extracing features, then worked upon by state-of-the-art custom built classifiers applied to a database consisting of gun type equipments. The key innovation is the explicit reformulation of different layers as residual functions, referencing the inputs given to different layers. The system is capable of real-time detection and demonstrates robustness across variations in affine transformations, scale, rotation, and partial closure or occlusion. The system undergoes cross-validation with different parameter values, and the optimal set of parameters for visual handheld gun detection is identified. The evaluation on the ImageNet dataset includes residual networks with a depth of up to 152 layers, which is 8 times deeper than VGG nets, yet achieves higher accuracy. The overall system performance is evaluated through Receiver Operating Characteristic (ROC) curves, illustrating the tradeoff between specificity and sensitivity.

**2.3.** *Integration with CCTV*

Further, Cheng [9] addresses the need for automated violence detection in surveillance camera footage, emphasizing the current prevalence of such cameras in public places. While these cameras have contributed to a reduction in the overall crime rate, their primary use has been for post-event analysis rather than real-time prevention. This model incorporates a unique feature by introducing a branch of the optical flow channel to contribute to the development of a pooling mechanism. They have introduced a new dataset and method for violence detection in surveillance videos. The RWF-2000 dataset is highlighted as the largest surveillance video dataset for violence detection in realistic scenes to date.
Additionally, [17] there was a need to discriminate between small hand-held objects such as smartphones from weapons like knifes, guns, etc for accurate prediction of violent action and efficient categorization of video feed captured from CCTVs. The focus is on detecting weapons and objects that could be mistaken for a handgun or knife when manipulated by hand in video surveillance scenarios.
The experimental study conducted using database containing six objects (pistol, knife, smartphone, bill, purse, and card) demonstrates that the proposed ODeBiC methodology effectively reduces the number of false positives through binarization techniques.

### III. METHODOLOGY

The system proposed is comprised of a Raspberry Pi, which is equipped with a camera module for the purpose of capturing real-time video data. The architecture of the system is separated into three primary components: the collection of images, image preprocessing, and image recognition, followed by the dissemination of results.

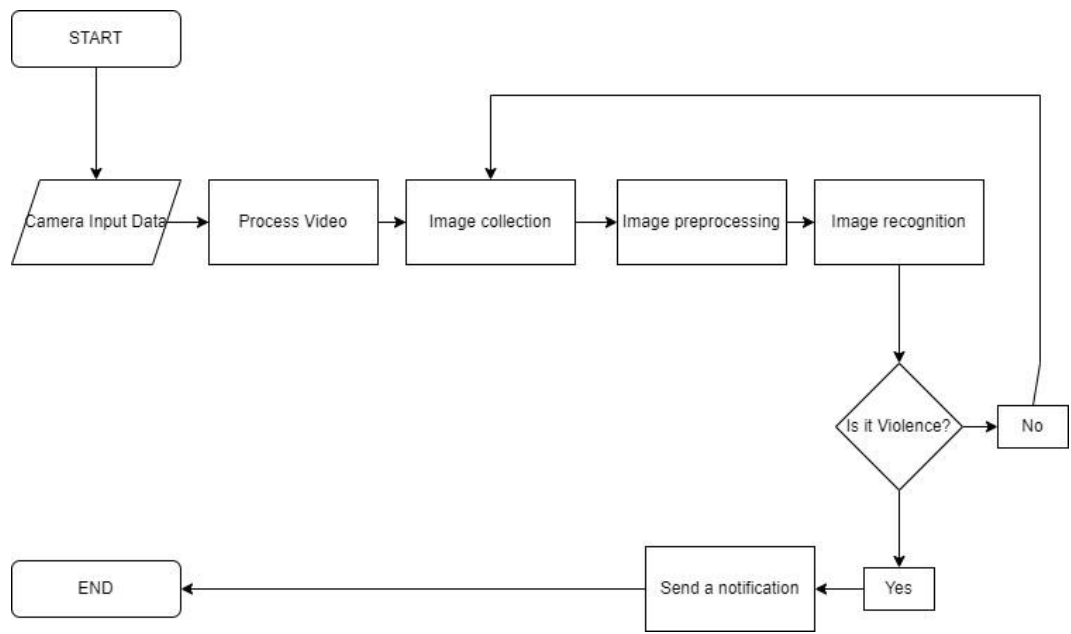

Fig 1: System Architecture Diagram

### 3.1. *Hardware Setup for Video capture*

The physical setup includes configuring the camera module with Raspberry pi to capture live video feed from surroundings where it is placed. The camera module is prepared by integrating it with a fisheye lens in order to attain maximum coverage of surroundings and attain wide area visibility.

Configuring a camera module with a Raspberry Pi involves attaching the module to the Pi's CSI port and integrating a fisheye lens for wide-area visibility. After installing the Raspberry Pi OS and camera software packages like `raspistill` and `raspivid`, additional streaming software may be installed for live streaming. Configuration includes adjusting camera settings such as resolution and exposure, calibrating for fisheye lens distortion, and setting up features like motion detection. Testing ensures functionality, while optimization involves adjusting settings for quality and performance, considering factors like lighting and placement. Regular monitoring and maintenance, including cleaning the lens and keeping software updated, help ensure continued operation and security of the setup. This process enables effective live video capture with wide area coverage using the Raspberry Pi and camera module.

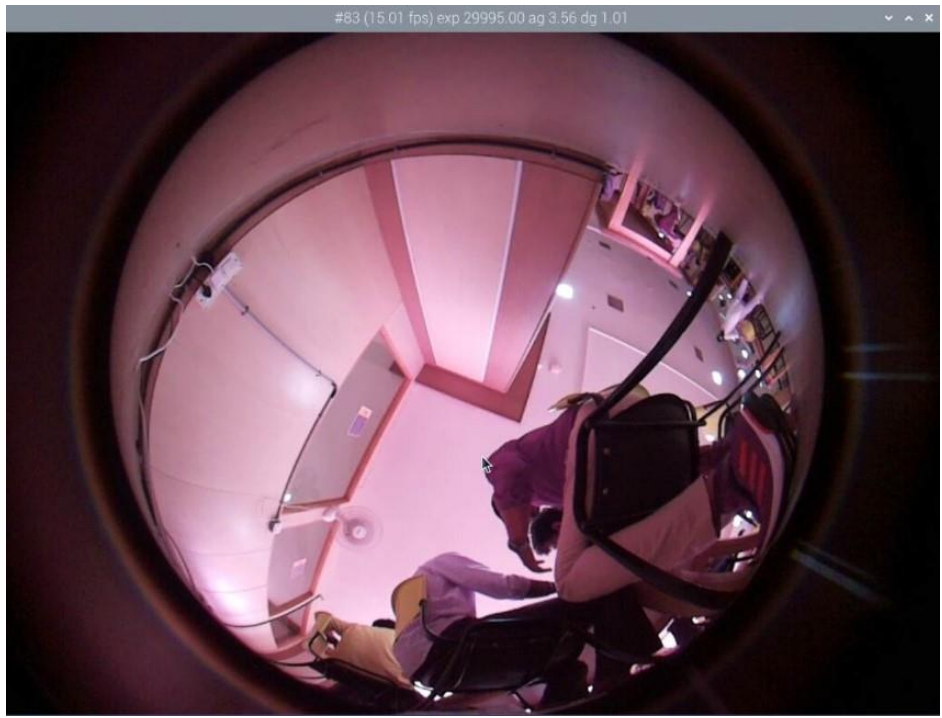

Fig 2: Sample image from camera sensor

### 3.2. *Preparation of Custom Dataset*

To train and enhance the performance of our ML model, we have prepared a custom dataset comprising a variety of videos. It has videos containing weapons and non-weapons for training and evaluation. It is mainly prepared by sampling some famous datasets such as the NTU CCTV fights, Hockey fights, Sohas and WVD datasets which have been used for violence detection models earlier. The videos are of 720p resolution with frame dimensions of 1280x720. A maximum of 500 videos are taken into account with length of 5 to 10 seconds per video for extraction of features

### 3.3. Training ML model

In order to develop our machine learning model for the detection of violence and the analysis of crime scenes, we utilize a CNN-LSTM algorithm that involves several critical stages. Initially, we have gathered our custom IIS (Intelligent Image Sensing) dataset, which consists of labeled video clips or frames indicating violent and non-violent activities. Next, we preprocess this data by extracting frames, resizing them uniformly, and converting them into a suitable format, such as numpy arrays. Following this, we employ a pre-trained CNN model, such as VGG or ResNet, to extract spatial features from each frame or train a CNN from scratch if the dataset size permits. Subsequently, we integrate an LSTM network to capture the temporal dynamics of the video sequence, inputting the sequence of feature vectors extracted by the CNN. After splitting the dataset into training, validation, and testing sets, we train the CNN-LSTM model, adjusting the hyperparameters to prevent overfitting based on the validation set performance. Finally, we assess the trained ML model using the custom testing dataset, utilizing parameters like accuracy, recall, and F1-score to determine its ability to detect violence.

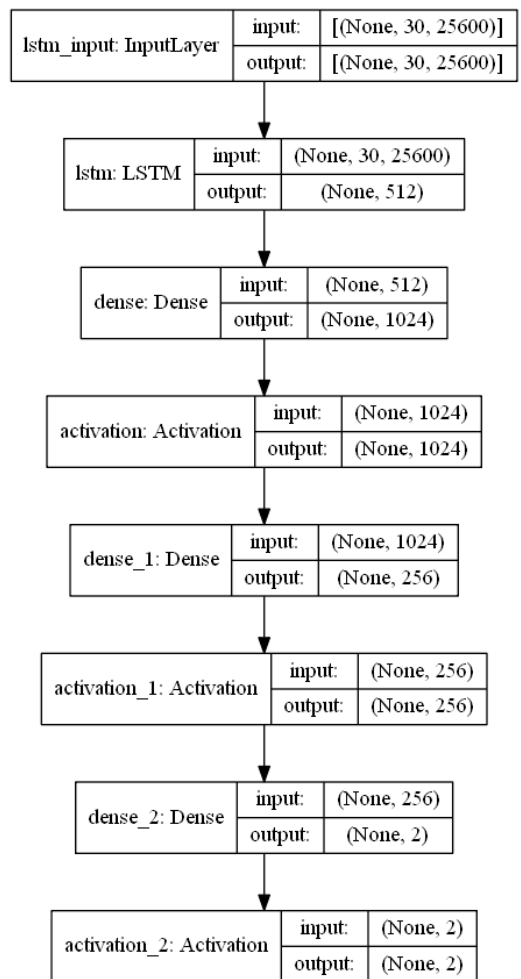

Fig 3: Model Architecture

### 3.4. Building up of Super Image

To construct a SUPER IMAGE, the primary concept revolves around preserving aspect information and minimizing information loss. This involves selecting a sample size, custom sampler, ratio of its sizes in different dimensions and spatial arrangement for rearrangement of frames meanwhile retaining the aspect ratio of the actual frames. For instance, in the case of 1280 × 720 frame resolution, where dimensions are in the form of height x width, given that width is greater than height, the aim is to produce a final constructed image with the height proximal to the width. This process ensures that the resulting image maintains the original aspect ratio while enhancing saliency and minimizing distortion. Through meticulous selection of sample sizes, samplers, and spatial arrangements, the approach maximizes the preservation of aspect information, ultimately yielding a SUPER IMAGE with optimized visual quality.

**Selection of Sampler**

The tdifferent types of samplers used is definitely a major factor in determining the ability of the various classifiers. To create a SUPER IMAGE, different sampling methods can be employed, each with its unique approach:

1. Uniform Sampler: This method selects frames which are uniformly distributed throughout the video by specifying the parameter k, which determines the number of frames to select. It calculates the stride between selected frames and then chooses k equally spaced indices. This ensures that frames are selected at regular intervals throughout the video.

2. Random Sampler: In contrast, the random sampler selects k frames randomly from the video without replacement. It utilizes the `np.random.choice()` function to generate a list of k unique random indices. This method introduces randomness in frame selection, ensuring a diverse representation of frames from the video.

3. Continuous Sampler: The continuous sampler chooses k frames evenly spread throughout the whole video. Similar to the uniform sampler, it calculates the stride between selected frames and then chooses k indices. However, this method ensures that frames are evenly distributed across the video timeline, providing a comprehensive representation of the video content over time.

By employing these sampling methods, the SUPER IMAGE construction process can effectively capture the salient aspects of the video while maintaining the desired aspect ratio and minimizing information loss.

4. Mean Absolute Difference (MAD) Sampler: In this approach, k frames are selected based on the smallest average absolute difference between adjacent frames. It calculates the absolute differences between each pair of adjacent frames and then selects the k frames with the smallest average absolute difference. This method aims to capture frames that exhibit minimal change from one frame to the next, ensuring consistency and stability in the selected frames.

5. Lucas-Kanade Sampler: Utilizing the Lucas-Kanade algorithm, this sampler computes optical flow between adjacent frames to measure motion. It selects k frames with the largest amount of motion, ensuring dynamic content is represented. By computing optical flow between adjacent frames and selecting frames with significant motion, this method captures the key dynamic elements of the video, providing a dynamic representation in the SUPER IMAGE.

These methods offer diverse approaches to frame selection, each suited to capture different aspects of the video content, such as temporal distribution, visual consistency, and motion dynamics. Integrating these samplers in the SUPER IMAGE construction process ensures comprehensive coverage and representation of the video content while maintaining the desired aspect ratio and minimizing information loss.

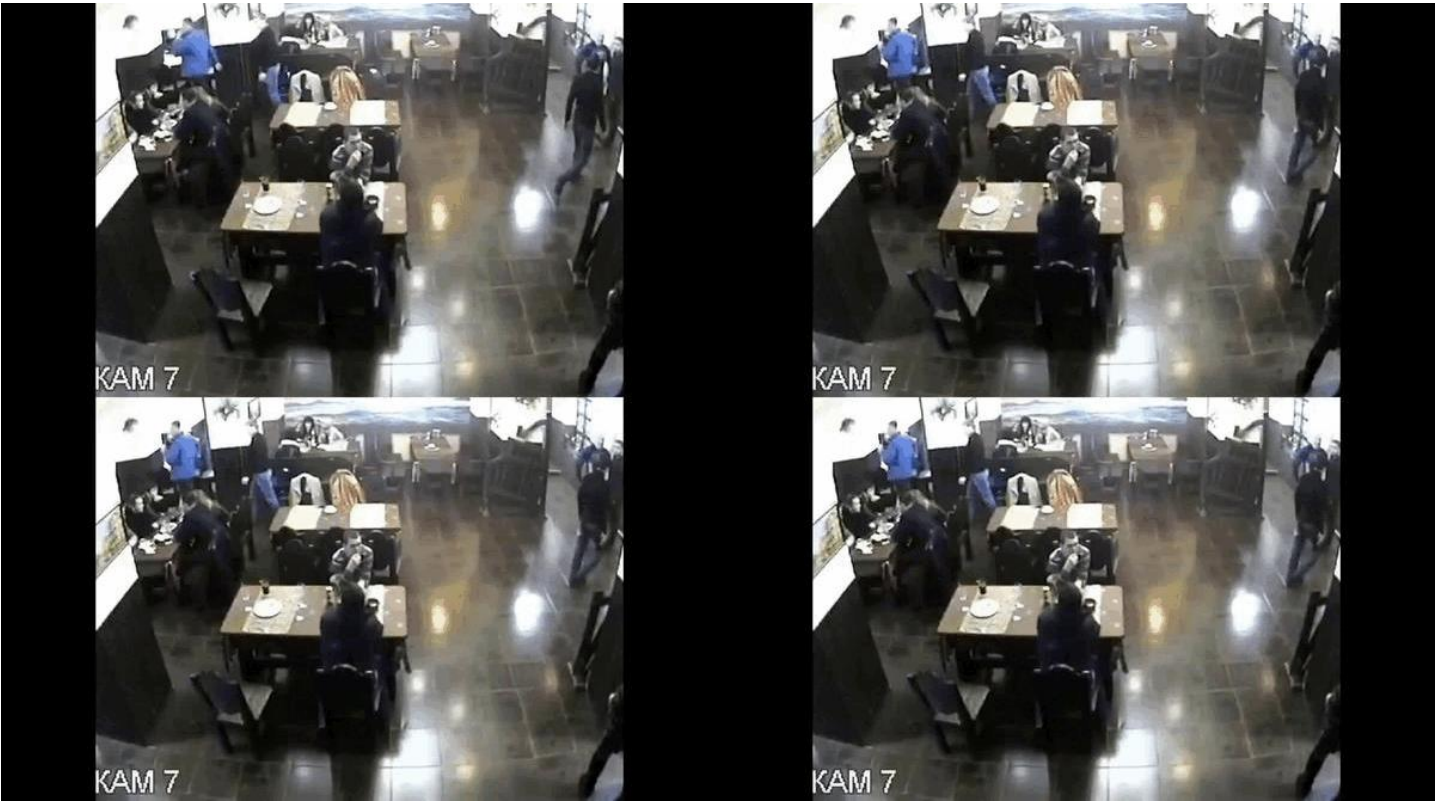

Fig 4: Sample Super Image

### 3.5. *Integration of Camera module with ML model*

Upon satisfactory performance, we need to deploy the model in a production environment and integrate it with hardware setups using frameworks like Flask for real-time violence detection. To do so, we have first set up Flask by installing it via pip and creating a new Python file for the Flask application. Within this application, we have defined routes to handle different types of requests, such as GET and POST. These routes contain logic to interact with both the hardware setup and the machine learning model. For hardware interaction, functions are called to perform tasks such as capturing images using the Raspberry Pi camera module. The machine learning model is integrated by loading it into the Flask application, either as a pre-trained model or one trained within the application. Routes or endpoints are created to send data to the model and receive predictions. Incoming requests from clients, such as web browsers or mobile apps, are handled within the Flask routes, processing data from the hardware and sending it to the model for prediction. The Flask server then returns appropriate responses, which may include predictions from the machine learning model or status updates about the hardware setup. Finally, the Flask server is run, ensuring accessibility to the hardware setup and any client applications needing interaction. This setup enables seamless real-time communication between the machine learning model and the hardware setup, facilitating a wide range of crime scene investigations. We will continuously monitor and improve the model, periodically retraining it on new data to ensure effectiveness in detecting violence over time. Through these steps, a robust CNN-LSTM model for violence detection can be developed and deployed for real-world applications.

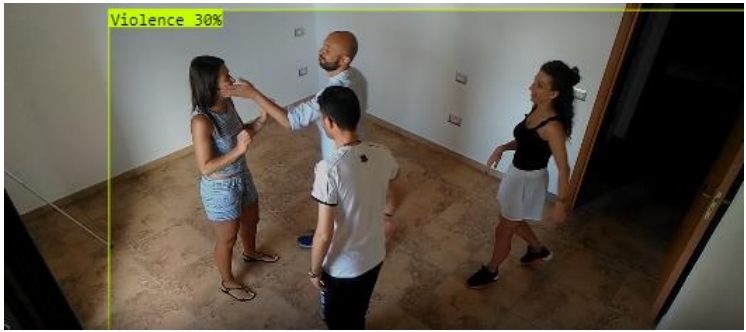

Fig 5: Model Output on Video

## IV. RESULTS

### 4.1. *Evaluation of Model Performance*

The performance of the Convolutional Neural Network (CNN) model was thoroughly assessed using various metrics, including accuracy, precision, recall, and F1-score. The system achieved a remarkable 83% accuracy in detecting violent events in real-time video streams. With precision and recall values at 91% and 74%, respectively, the model effectively identified true positive cases of violence without a significant number of false positives. The F1-score, which consolidates precision and recall, was calculated to be 90%, showcasing the balanced performance of the detection system.

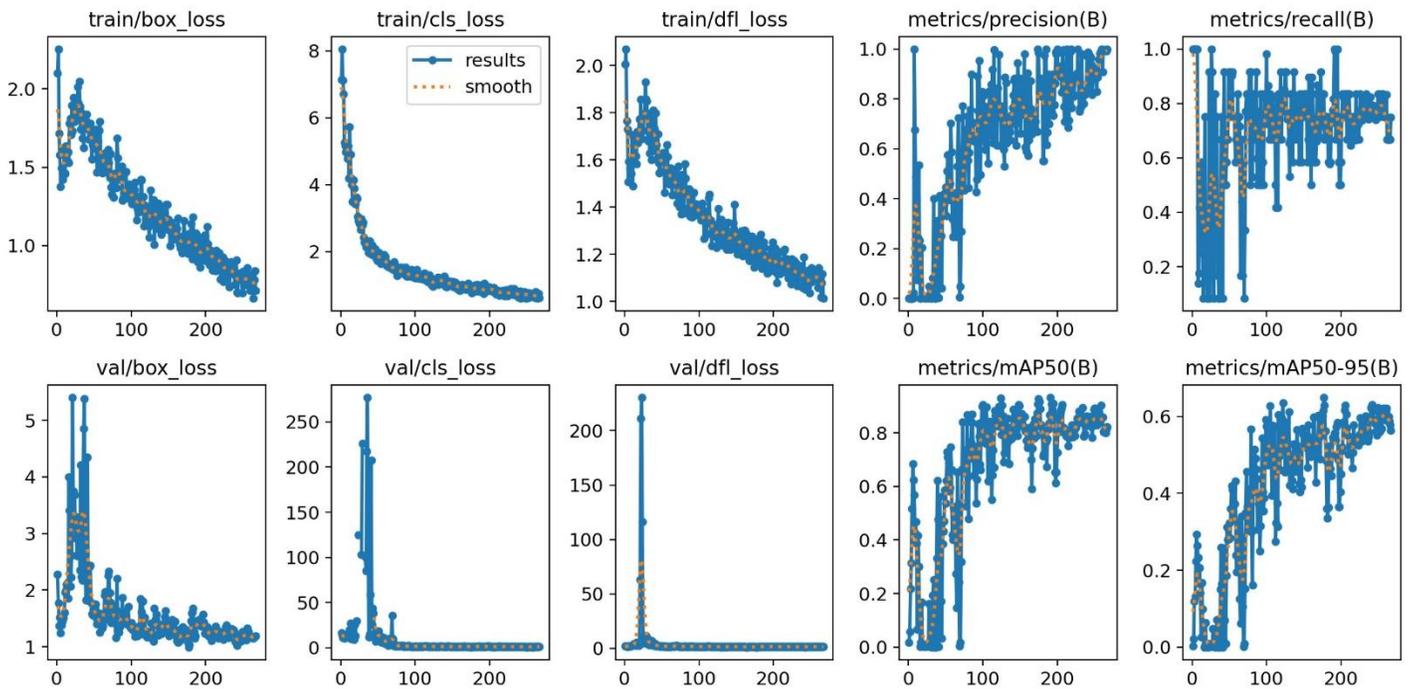

Fig 6: Graphical Evaulation of Model Performance

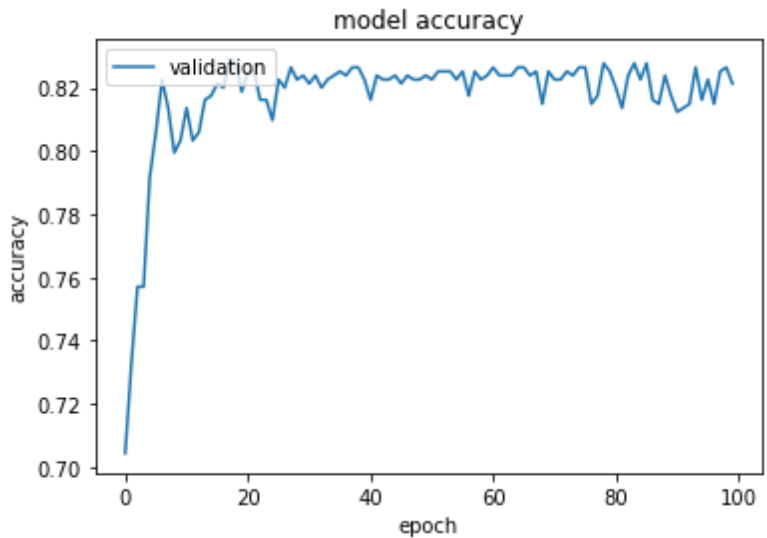
Fig 7: Accuracy Chart of Model

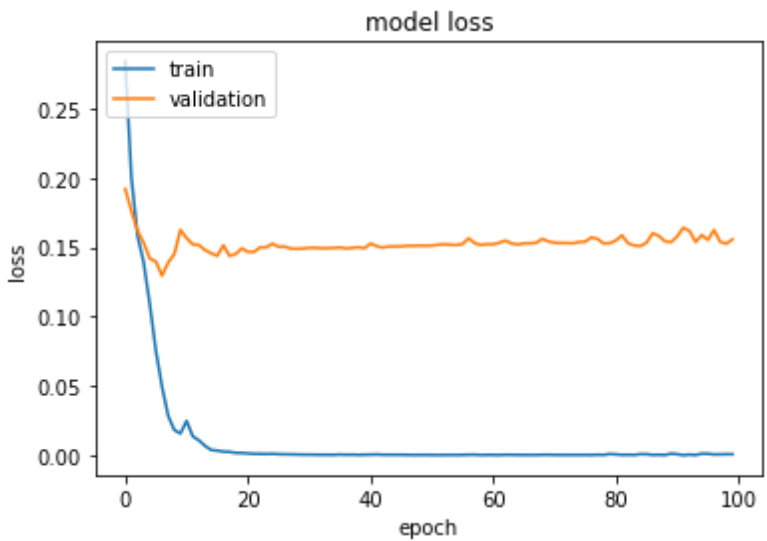
Fig 8: Loss Chart of Model

4.2. *Real-World Testing*

To evaluate the system's real-world effectiveness, field tests were conducted in various environments with different levels of activity and lighting conditions. These tests revealed the system's resilience and adaptability, with consistent performance metrics across the different settings. During these tests, the system successfully flagged incidents of violence, enabling prompt responses.

4.3. *Computational Efficiency*

The Raspberry Pi's computational efficiency was a focal point of analysis. The device processed real-time data effectively, with an average processing time of 2 seconds per frame. This performance, considering the hardware limitations of the Raspberry Pi, highlights the optimization of the ML model and the overall system design.

4.4. *System Limitations and Challenges*
Despite the promising results, certain challenges were identified, particularly in high-density crowd scenarios and low-light conditions. Under these complex conditions, accuracy slightly decreased, indicating areas for future improvement.

# V.    CONCLUSION

This study has demonstrated the feasibility and effectiveness of using a machine learning-based image sensing system on a Raspberry Pi to detect violent incidents in real-time video streams. The employment of convolutional neural networks (CNNs) on a cost-effective platform like the Raspberry Pi presents a promising solution for enhancing public safety through advanced surveillance capabilities.

The results show that the system attains high levels of accuracy, precision, recall, and F1-scores, making it a viable tool for real-world applications in violence detection. The system's performance in diverse environmental settings during field tests further validates its robustness and adaptability, establishing its potential for broader deployment. Despite facing challenges in high-density and low-light conditions, the system maintained commendable performance levels, underscoring the effectiveness of the CNN model and the preprocessing techniques utilized. Moreover, the implementation on Raspberry Pi highlights the potential for widespread application of this technology, particularly in regions with limited access to advanced computational resources.

It proves that sophisticated machine learning models can be effectively executed on low-cost hardware, making advanced surveillance technology more accessible and cost-effective. Future work will focus on addressing the identified limitations by refining the model and incorporating additional sensors for better performance in challenging scenarios. Further research into enhancing privacy and ethical practices is also crucial as the technology progresses and becomes more integrated into public security systems. In conclusion, this research contributes to the advancing field of crime analysis and public safety, paving the way for more intelligent, efficient, and responsive surveillance systems that leverage the power of machine learning and the accessibility of consumer-grade technology.